\documentclass[letterpaper]{article} 
\pdfoutput=1
\usepackage[preprint]{aaai2027}  
\usepackage[hyphens]{url}  
\usepackage{graphicx} 
\usepackage{natbib}  
\usepackage{caption} 
\usepackage{booktabs}
\usepackage{amsmath}
\usepackage{amssymb}
\newcommand{\method}{VESPA}
\newcommand{\methodexpanded}{\textbf{VE}rifiable \textbf{S}ymbolic \textbf{P}olicy \textbf{A}daptation}
\newcommand{\baseline}{\textsc{Uninformed}}

\title{\method{}: Verifiable Symbolic Policy Adaptation for Safe Reinforcement Learning}
\author{
    Paul K\"onig\thanks{Corresponding author: paul.koenig@tu-clausthal.de}\textsuperscript{\rm 1},
    Sascha Marton\textsuperscript{\rm 1},
    Stefan L\"udtke\thanks{Equal supervision.}\textsuperscript{\rm 2},
    Christian Bartelt\footnotemark[2]\textsuperscript{\rm 1}
}
\affiliations{
    \textsuperscript{\rm 1}COgnitive SoftwaRE (CORE), Clausthal University of Technology\\
    \textsuperscript{\rm 2}Institute for Visual and Analytic Computing, University of Rostock
}

\begin{document}

\maketitle

\begin{abstract}
Safe reinforcement learning (RL) aims to prevent agents from taking actions that are dangerous or destructive to their environment. This is especially important for safety-critical domains, such as robotics or autonomous driving. Formal verification can be used to prove agent safety through safety constraints. However, in practice, many policies do violate safety constraints, and thus are verified to be unsafe. Since the goal is to obtain a safe policy, we would need to retrain a new policy from scratch and then verify it again until a candidate happens to pass. Both training and verification are usually not cheap, which makes this process inefficient. We introduce \method{} (\methodexpanded{}), a framework for targeted fine-tuning of decision tree policies. We show that using counterexamples for policy adaptation can make existing unsafe policies safe, or at least make considerable progress towards safety. Targeted fine-tuning of the existing policy on those counterexamples enables it to learn how to avoid violating safety constraints. We thus use the result of a formal verification as a training signal for fine-tuning. On CartPole, \method{} raises the mean number of verified safe steps from 63.8 to at least 92.7, reaching after two iterations what the same loop without counterexamples reaches after ten. On MountainCarContinuous, \method{} turns an unsafe policy into a safe policy in 53 of 60 runs. We conclude that counterexamples from a verifier are an effective training signal, which can be utilized to systematically improve the safety of policies. Our implementation is available at: \url{https://github.com/p-koenig/VESPA}
\end{abstract}

\section{Introduction}
\label{sec:intro}
Reinforcement learning (RL) has recently achieved remarkable progress in the areas of robotics and control \citep{tang2025deep}. Generally, both the training and evaluation of RL algorithms are based on expected return. However, solely optimizing the expected return for a given task is inadequate for safety-critical applications, because the environment or agent can impose safety constraints which must not be violated. For instance, in robotic applications it is insufficient to evaluate solely on average task performance. Instead, we need to guarantee that unsafe behaviors, such as a humanoid robot falling over, never occur. RL does not give such guarantees by default.

In the area of safe RL, numerous approaches have been proposed to overcome this problem. As discussed by \citet{zhao2023state}, some approaches provide proofs of safety, while others give probabilistic safety guarantees. We argue that any form of probabilistic safety guarantee is insufficient for deploying RL in safety-critical domains. In environments with high risks, a formal verification for the deployed policy is needed. Safety can either be enforced at runtime, during training or inference, as in shielding \citep{alshiekh2018shielding}, or verified offline as a one-time proof for a fixed policy. Reachability analysis \citep{althoff2021set} is a versatile approach to offline verification, and issues a final verdict (either \textit{safe} or \textit{unsafe}) once for a fixed policy.

One challenge of applying reachability analysis in general is the gap between verifying whether a policy is safe and obtaining a safe policy. Formal verification only returns the verdict \textit{safe} or \textit{unsafe}. Since the final goal is to create a safe policy, one must train and verify many independent candidates, which is inefficient and in some cases infeasible. 

To overcome this challenge, we introduce \method{} (\methodexpanded{}) and show how to fine-tune a given policy based on the result of a formal verification. We focus on symbolic tree-based policies \citep[SYMPOL,][]{marton2025sympol}, since their axis-aligned splits make them amenable to reachability analysis. For policies with an \textit{unsafe} verdict, we natively obtain counterexamples, i.e., states that are reachable by the policy and violate the safety constraints. Those are then traced back to some predecessor state region and used as start states for fine-tuning the policy. This targeted policy adaptation teaches the policy to steer away from the unsafe counterexamples. In this way, \method{} utilizes the output of the reachability analysis as a training signal to fine-tune the current policy and obtain the next candidate policy. More specifically, our contributions are as follows:
\begin{itemize}
    \item We present a verifier for tree policies that scales to longer horizons than other approaches (Section~\ref{sec:reach}).
    \item We utilize failures extracted by the verifier for targeted fine-tuning, the core of \method{} (Section~\ref{sec:pa}).
    \item We show that \method{} obtains fully safe policies on MountainCarContinuous most of the time and makes considerable progress towards safe CartPole policies (Section~\ref{sec:results}).
\end{itemize}

\begin{figure*}[t]
  \centering
  \includegraphics[width=\textwidth]{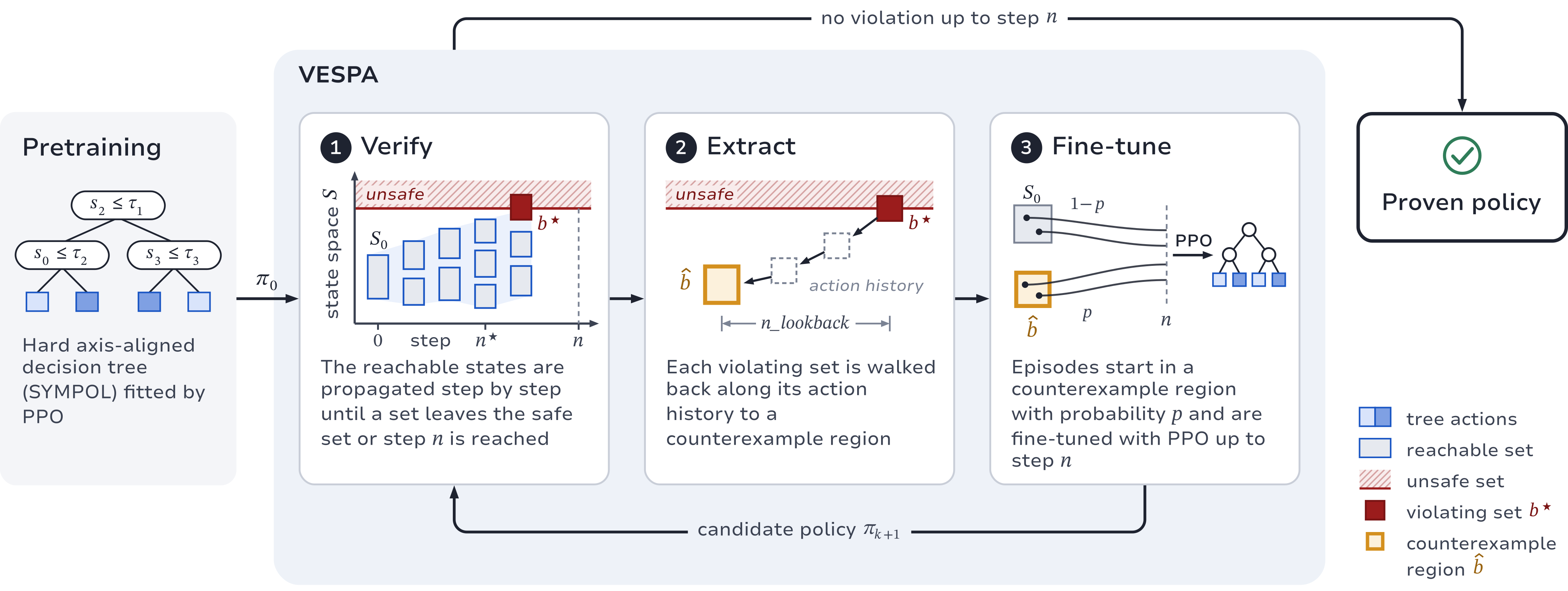}
  \caption{Overview of \method{}. A hard, axis-aligned SYMPOL decision tree policy $\pi_0$ is pretrained with PPO. \textbf{(1)~Verify:} starting from the start states $S_0$, the reachable states in the state space $S$ are propagated step by step until the first violating step $n^{\star}+1$ or the horizon $n$. A single verification step is detailed in Fig.~\ref{fig:verify}. \textbf{(2)~Extract:} each violating set $b^{\star}$ is walked back for \textit{n\_lookback} steps along its action history, giving a counterexample region $\hat{b}$. \textbf{(3)~Fine-tune:} episode start states are sampled from the counterexample regions with probability $p$ and from $S_0$ otherwise, and the tree is fine-tuned with PPO up to step $n$. The candidate policy $\pi_{k+1}$ is verified again. The loop ends with a proof of Eq.~\ref{eq:safety-property} once no violation occurs up to step $n$.}
  \label{fig:overview}
\end{figure*}

On CartPole, our verifier scales to a mean horizon of at least 92.7 steps, more than the 10 steps verified by prior work on tree policies for CartPole \citep{bastani2018viper,vasic2022moet}. Our policy adaptation component then increases the proven steps by a mean of at least 29 steps.

On MountainCarContinuous, we prove in 53 of 60 runs that \method{} turns unsafe policies into safe policies for the full 200-step horizon from all start positions defined by the environment \citep{lange2022gymnax}, whereas related proofs start from only 15--30\% of these start positions \citep{schilling2023dtcs,huang2022polar}. Our analysis also shows that expected return is a poor proxy for the safety of RL policies, even if the reward signal covers the safety constraint completely: On CartPole, none of the tested policies with near-maximal return is verified safe for 200 steps, and one is verified for only 27 steps.

\section{\method{}: \methodexpanded{}}
\label{sec:vespa}

We introduce \method{}, an alternating verification and fine-tuning loop (Fig.~\ref{fig:overview}) based on three main components:
\begin{itemize}
    \item \textbf{Verification} for symbolic tree-based policies through reachability analysis over a continuous state space (Section~\ref{sec:reach})
    \item   \textbf{Counterexample extraction} based on backtracking of unsafe state regions (Section~\ref{sec:cex})
    \item \textbf{Policy adaptation} through mixing original start states and counterexample start states and fine-tuning on the remaining horizon (Section~\ref{sec:pa})
\end{itemize}

\subsection{Background}
\label{sec:background}
Following \citet{zhao2023state}, we distinguish between state-wise and cumulative safety constraints. State-wise constraints have to be met at each step, while cumulative constraints have to be met for an entire episode taking all steps into account. Within the former, we focus on \textit{invariance} constraints, which force the agent to never leave a constraint-satisfying state region.  Within the latter, we focus on \textit{reachability} constraints, which describe a goal which has to be met at some step of the episode.

\paragraph{Reachability Analysis}
This approach to formal verification \citep{althoff2021set}, here applied to RL policies, identifies the set of reachable states, which can then be checked against some safety constraint. The set of reachable states depends on the set of start states $S_0$ and the environment transition function $f$, as well as the policy $\pi(s)$. Starting from $S_0$, the set of reachable states $S_i$ at each step $i$ is determined by:
\[
S_{i+1} \;=\; \bigl\{\, f\bigl(s, \pi(s)\bigr) \;\big|\; s \in S_i \,\bigr\},
\qquad i = 0, \dots, n-1.
\]

\paragraph{Tree-Based Policies}
The model type has great influence on the performance and interpretability of RL policies. While Deep Q-Network (DQN) policies and other neural network policies received much attention in recent years, tree-based policies \citep{marton2025sympol,fuhrer2025gradient} achieved competitive results on several benchmarks. In addition to their better interpretability \cite{marton2025sympol}, tree-based policies naturally provide hard axis-aligned decision boundaries. We exploit this property within our verification algorithm, to make verification more efficient.

\subsection{Reachability Analysis with Tree Policies}
\label{sec:reach}
\begin{figure*}[t]
  \centering
  \includegraphics[width=\textwidth]{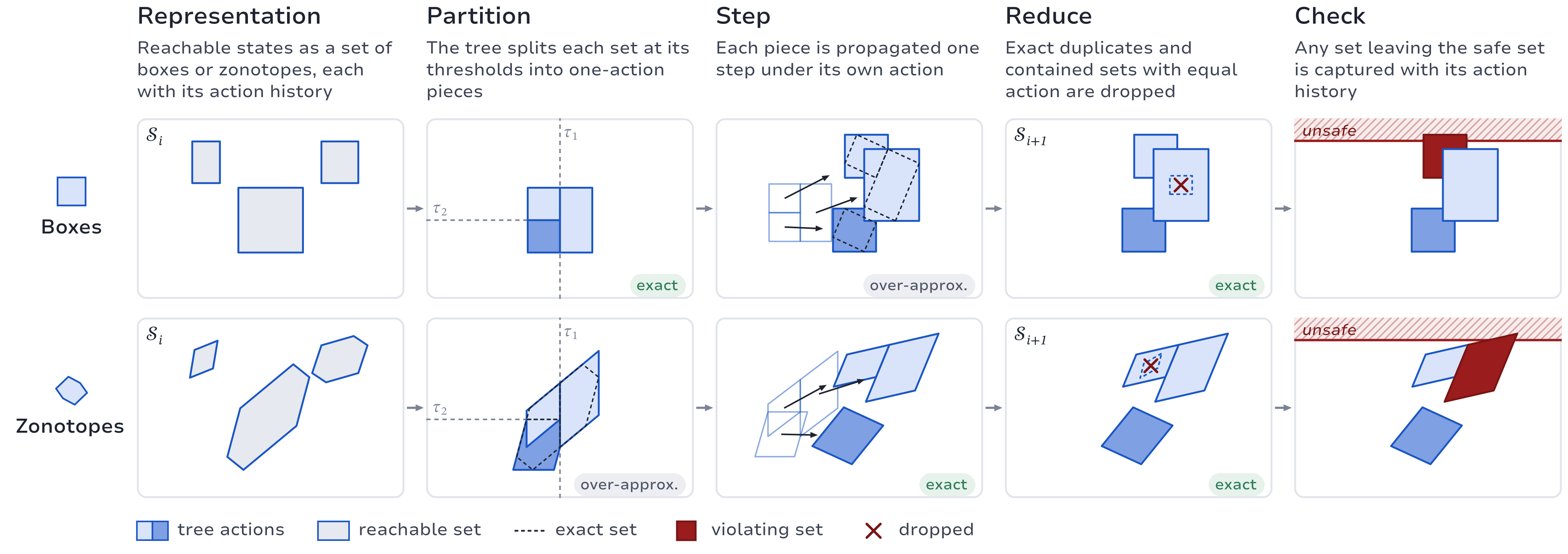}
  \caption{One verification step of \method{}. Both rows show the same step from $\mathcal{S}_i$ to $\mathcal{S}_{i+1}$, with boxes (top) and zonotopes (bottom) as state representation. \textbf{Representation:} the reachable states $\mathcal{S}_i$ are represented as a set of boxes or zonotopes, each carrying its action history. \textbf{Partition:} the axis-aligned splits of the policy divide each set into pieces, each with a single action $a_j$. This is exact for boxes, while each zonotope piece is over-approximated by an enclosing zonotope. \textbf{Step:} each piece is propagated through $\hat{f}(\cdot, a_j)$. For boxes, interval arithmetic over-approximates by an enclosing box, while affine transitions of zonotopes are exact. \textbf{Reduce:} exact duplicates and sets contained in another set with equal action are dropped (Eq.~\ref{eq:reduce}), giving $\mathcal{S}_{i+1}$ (Eq.~\ref{eq:reach-step}). \textbf{Check:} every set that leaves $S_{\mathit{safe}}$ is captured as $b^{\star}$ with its action history at the first violating step $n^{\star}+1$ (Eq.~\ref{eq:first-unsafe}). Otherwise, $\mathcal{S}_{i+1}$ becomes the next frontier, and after step $n$ the verification returns a proof of Eq.~\ref{eq:safety-property}.}
  \label{fig:verify}
\end{figure*}
\method{} verifies policies with set-based reachability analysis \citep{althoff2021set}, which has also been applied to tree policies before \citep{schilling2023dtcs}: sets of states are propagated step by step through the closed loop of policy and environment and checked against the property. In our case, the policy acts at discrete time steps and selects one action per axis-aligned leaf, so the tree splits divide each set into pieces with a single action each (Fig.~\ref{fig:verify}).

We consider a Markov Decision Process with state space $S \subseteq \mathbb{R}^{d}$, action set $A$, deterministic transition function $f$ and initial states $S_{0} \subseteq S$. We assume that a pretrained axis-aligned decision tree policy $\pi$ already exists.  \method{} does not make any assumptions on how this policy was obtained. Given a safe set $S_{\mathit{safe}}$ (or a goal set $S_{\mathit{goal}}$) and a verification horizon $n$, a safe policy is defined as \cite{baier2008principles,biere1999symbolic}:
\begin{equation}
\label{eq:safety-property}
\forall s_{0} \in S_{0}\; \forall t \le n : \; s_{t} \in S_{\mathit{safe}},
\end{equation}
for \emph{invariance} properties, and:
\begin{equation}
\label{eq:reachability-property}
\forall s_{0} \in S_{0}\; \exists t \le n : \; s_{t} \in S_{\mathit{goal}},
\end{equation}
for \emph{bounded reachability} properties, where $s_{t}$ is the state reached after $t$ policy steps from $s_{0}$.

\paragraph{Exploiting Hard Axis-Aligned Decision Boundaries}
As a default, we use $d$-dimensional boxes to represent a subset region of a continuous state space. A decision tree policy with $L$ leaves induces $L$ subset regions, one for each leaf. Each region corresponds to a single action that is selected for that region. We then intersect the start state region $S_0$ with every leaf region and obtain at most $L$ parts of $S_0$. We apply the transition function to each part individually, obtaining the results of one policy step for all start states $S_{0}$. This process is repeated for each resulting box recursively along the verification horizon $n$ (Fig.~\ref{fig:verify}). Formally, let $R_1, \dots, R_L$ be the axis-aligned leaf regions of the tree and $a_j$ the action of leaf $j$. Starting from $\mathcal{S}_0$, one policy step maps the set of boxes $\mathcal{S}_i$ to
\begin{equation}
\label{eq:reach-step}
\mathcal{S}_{i+1} = \mathrm{Reduce}\Big( \bigcup_{b \in \mathcal{S}_i} \bigcup_{j=1}^{L} \big\{ \hat{f}(b \cap R_j,\, a_j) \big\} \Big),
\end{equation}
where $\hat{f}$ is the over-approximated environment step described below and $\mathrm{Reduce}$ applies the reduction techniques in Eq.~\ref{eq:reduce}.

For \textit{invariance} properties, each box in $\mathcal{S}_i$ is intersected with the complement of the safe set, $\overline{S_{\mathit{safe}}}$. The verdict is \textit{unsafe} if this intersection is non-empty, i.e., some box leaves the safe set, with the number of verified safe steps
\begin{equation}
\label{eq:first-unsafe}
n^{\star} = \min \{\, i \le n \mid \exists\, b \in \mathcal{S}_i : b \cap \overline{S_{\mathit{safe}}} \neq \emptyset \,\} - 1,
\end{equation}
and \textit{safe} otherwise. The boxes $b^{\star} \in \mathcal{S}_{n^{\star}+1}$ with $b^{\star} \cap \overline{S_{\mathit{safe}}} \neq \emptyset$ are the counterexamples (Section~\ref{sec:cex}). 

Naturally, $n^{\star}$ only exists for \textit{invariance} properties. For \textit{reachability} properties, the reachable set is intersected with the complement of the goal set, $\overline{S_{\mathit{goal}}}$, at every step. A box that lies entirely inside $S_{\mathit{goal}}$ has satisfied Eq.~\ref{eq:reachability-property} and is absorbed. Only the remaining boxes
\begin{equation}
\label{eq:goal-open}
\mathcal{S}^{\circ}_i = \{\, b \in \mathcal{S}_i \mid b \cap \overline{S_{\mathit{goal}}} \neq \emptyset \,\}
\end{equation}
are propagated, i.e., Eq.~\ref{eq:reach-step} is applied to $\mathcal{S}^{\circ}_i$ in place of $\mathcal{S}_i$. The verdict is \textit{unsafe} if some box has still not reached the goal at the horizon, $\mathcal{S}^{\circ}_n \neq \emptyset$, and \textit{safe} otherwise. The boxes in $\mathcal{S}^{\circ}_n$ are the counterexamples $b^{\star}$ (Section~\ref{sec:cex}). In both cases, \textit{invariance} and \textit{reachability}, verifying the constraint is an efficient intersection, done in constant time per box.
Since the leaf regions are axis-aligned, the intersections $b \cap R_j$ are exact. We over-approximate the environment dynamics $f$ with $\hat{f}$ by interval arithmetic \cite{moore2009interval}. A \textit{safe} verdict is always a proof, while an \textit{unsafe} verdict may be inaccurate. Combining these properties with Counterexample Extraction (Section~\ref{sec:cex}) and Policy Adaptation (Section~\ref{sec:pa}), we are able to fine-tune and then prove a \textit{safe} policy.

\paragraph{Zonotopes as an Alternative State Representation}
Instead of boxes, \method{} can represent each element of $\mathcal{S}_i$ as a zonotope \cite{girard2005zonotopes}, i.e., a centrally symmetric set with center $c \in \mathbb{R}^{d}$ and generator matrix $G \in \mathbb{R}^{d \times m}$. We use zonotopes as an alternative representation of the observed state space: Eq.~\ref{eq:reach-step}, the counterexample extraction and the policy adaptation step remain unchanged. Their main advantage is that affine state transitions are exact, so correlations between state dimensions are preserved and the transitions do not need to be over-approximated as with boxes. However, the axis-aligned tree splits are no longer exact, since $Z \cap R_j$ is in general not a zonotope and has to be over-approximated \cite{girard2008hyperplane}. Which representation is tighter depends on the environment: zonotopes are preferable when the environment dynamics strongly couple state dimensions, boxes when the fragmentation by the tree dominates (Appendix~\ref{sec:app-verify}). As with boxes, a \textit{safe} verdict obtained with zonotopes is a proof, while an \textit{unsafe} verdict may be inaccurate.

\paragraph{Scaling to Many Environment Steps}
The computational effort of verification within \method{} depends on two main factors: the structure of the pretrained policy, with larger trees leading to more fragmentation, and the number of observed dimensions as defined by the environment. The computational effort also grows with higher verification depth, i.e., more verified steps. Most environments define a step limit, after which an episode is either terminated or truncated.

To reduce the computational effort, we introduce two reduction techniques aimed at reducing the fragmentation of the reachable state space. First, exact duplicates are dropped. Second, if a state region is contained in another region, it can be safely dropped. In both cases, their corresponding actions need to match.
Formally, with $a(b)$ the action under which zonotope or box $b$ was propagated in the current step, and $\mathcal{S}$ the set of zonotopes or boxes after dropping exact duplicates,
\begin{equation}
\label{eq:reduce}
\begin{aligned}
\mathrm{Reduce}(\mathcal{S}) &= \{\, b \in \mathcal{S} \mid \nexists\, b' \in \mathcal{S} \setminus \{b\} : \\
&\qquad a(b') = a(b),\; b \subseteq b' \,\}.
\end{aligned}
\end{equation}
Both reduction techniques are used for both box and zonotope representations. Applying those reduction techniques is lossless and does not impact the verdict.

\subsection{Counterexample Extraction}
\label{sec:cex}
An \textit{unsafe} verification verdict comes with at least one violating set $b^{\star}$, i.e., a reachable box or zonotope that leaves $S_{\mathit{safe}}$ at the first violating step $n^{\star}+1$. We then extract the predecessor state regions of every violating set.

\paragraph{Extracting Predecessor State Regions}
Naively using the violating set itself as a counterexample for policy adaptation would be a poor choice. Since the violation has already occurred, it would be impossible to recover. The counterexample extraction of \method{} therefore returns a part of some predecessor set, which we call a counterexample region $\hat{b}$. The set of all extracted counterexample regions is $\hat{\mathcal{B}}$. To accomplish this, we keep a record of the action history for each set, which we then use to backtrack its predecessors. The length of the action history, and thus the number of backtracked steps, is denoted by \textit{n\_lookback}. Starting from only the violating part of each violating box $b^{\star} \setminus S_{\mathit{safe}}$, we apply the inverse transition $\hat{f}^{-1}$ once per recorded action, from the most recent to the oldest, each time obtaining the box one step earlier. The appropriate value of \textit{n\_lookback} depends on the property class. \textit{Reachability} properties naturally require substantially larger values, since a failure to reach the goal is only detected at the horizon and may originate at any earlier step.

\subsection{Policy Adaptation}
\label{sec:pa}
The \method{} verification issues either a safety proof or counterexamples, which are regions reachable by the current policy from which a future violation is inevitable.  Since the main goal is to train a safe policy, these counterexamples are valuable: they pinpoint states that lead to potentially very rare trajectories in which the constraint is violated. Standard RL training would encounter such states only by chance during exploration, and may never sample them at all. Without targeted guidance, finding a safe solution is hard. We therefore utilize the counterexamples for an informed policy adaptation, directing training exactly to these hard cases instead of relying on undirected exploration.

\paragraph{Initial-State Distribution} The initial-state distribution defines the sampling of training start states and thereby which episodes the policy is trained on \citep{barto2018reinforcement}. During standard pretraining, it is defined by the environment and usually covers the range of possible start states uniformly. In policy adaptation, we focus the fine-tuning on the hard regions that led to violations. Therefore, we modify the initial-state distribution to sample with probability $p$ from the extracted counterexample regions and with probability $1-p$ from the standard start state region, i.e.,
\begin{equation}
\label{eq:mixed-init}
\mathcal{D}_{\mathit{mix}} = (1-p)\,\mathcal{D}_0 + \frac{p}{|\hat{\mathcal{B}}|} \sum_{\hat{b} \in \hat{\mathcal{B}}} \mathcal{U}(\hat{b}),
\end{equation}
where $\mathcal{D}_0$ is the initial-state distribution of the environment and $\mathcal{U}(\hat{b})$ the uniform distribution over region $\hat{b}$. Every counterexample region is thus drawn with equal probability, independent of its volume. The mixture thus focuses on the hard edge cases while preventing forgetting \citep{florensa2017reverse,ivanovic2019barc}.

\paragraph{Standard Fine-Tuning for the Remaining Horizon} Our fine-tuning supports any on-policy optimization algorithm. In the case of actor-critic methods such as PPO \citep{schulman2017ppo}, we reinitialize the critic from scratch, since the training data distribution has changed. We then run the standard formulation of the optimizer, but truncate each episode at the end of the remaining horizon. Depending on the unsafe step and the \textit{n\_lookback} parameter, we may start a training episode in the middle of the horizon, far away from the start state region. Therefore, it is natural to run those episodes only for the remaining horizon.

\section{Evaluation}
\label{sec:exp}
We evaluate whether \method{} turns pretrained SYMPOL trees \cite{marton2025sympol} that fail verification into policies that are verified safe over the full horizon on two classic control benchmarks. Moreover, we show that the average return of a policy and its verified safe steps are decoupled.

\subsection{Experimental Setup}
\label{sec:setup}

\paragraph{Environments and Safety Constraints}
We carry out our evaluation on the CartPole (CP) \cite{barto1983cartpole} and MountainCarContinuous (MC-C) \cite{moore1990efficient} control environments (Appendix~\ref{sec:app-verify}). Solving CP requires the policy to balance a pole on a cart while controlling the movement of the cart. An episode fails if either the pole falls over or the cart exits the allowed area. Solving MC-C requires a car to gain energy from multiple swing-ups to escape a valley and finally reach the goal position at the top of the right hill.
For both environments, pretraining and the reported average return follow the standard formulation, while both constraints use the horizon $n = 200$, the episode limit of the original CartPole-v0 and MountainCar-v0 specifications \cite{brockman2016gym,verma2018pirl}.

\paragraph{Pretraining}
\method{} does not make any assumptions on how the pretrained axis-aligned tree policy was obtained. We focus on the SYMPOL configurations of \citet{marton2025sympol} for both CP and MC-C (Appendix~\ref{sec:app-pretrain}).

\paragraph{Pruning}
We apply the pruning of the SYMPOL implementation \cite{marton2025sympol} and evaluate both the pruned and the unpruned pretrained policies. Pruning does not change the actions of a policy, but the pruned tree has fewer nodes that the fine-tuning can adjust. We therefore treat the pruned and the unpruned tree of a pretraining seed as two separate runs, although they are not independent. We provide an analysis of pruned and unpruned policies in Appendix~\ref{sec:app-results}, which shows no evidence of any impact.

\paragraph{Verification and Counterexample Extraction}
We verify CP with boxes and MC-C with zonotopes (Section~\ref{sec:reach}, Appendix~\ref{sec:app-verify}). For MC-C, we additionally divide $S_0$ into $2^{14} = 16384$ start state regions of equal width along the position $x$ and verify each region separately. For the counterexample extraction (Section~\ref{sec:cex}), we use \textit{n\_lookback} $= 10$ for CP and \textit{n\_lookback} $= n = 200$ for MC-C (Appendix~\ref{sec:app-config}).

\paragraph{Policy Adaptation}
In both environments, fine-tuning uses the PPO hyperparameters of \citet{marton2025sympol} with few changes (Appendix~\ref{sec:app-config}), and we sample fine-tuning start states from the counterexample regions with probability $p = 0.5$ (Eq.~\ref{eq:mixed-init}).
Each run consists of at most 11 fine-tuning iterations for CP and at most 5 for MC-C. A candidate replaces the current policy only if it is verified at least as deep (CP), or if it is \textit{safe} or its ratio of safe start state regions is strictly higher (MC-C). We stop either after the verdict is \textit{safe}, after the last iteration, or, for CP, when the verification of a candidate runs out of memory beyond the depth of the current policy.

\paragraph{Metrics}
For CP, our main metric is the number of verified safe steps $n^{\star}$ (Eq.~\ref{eq:first-unsafe}). If the verification runs out of memory before a verdict, we report the steps verified so far as a lower bound on $n^{\star}$ (Appendix~\ref{sec:app-config}). For MC-C, we report the ratio of safe start state regions. A start state region is safe if all trajectories provably reach the goal within $n$ steps (Eq.~\ref{eq:reachability-property}). For both environments, we additionally report the number of runs in which \method{} obtains a \textit{safe} policy and, as a secondary metric, the average return over 30 episodes with greedy action selection (1000 episodes in Fig.~\ref{fig:return-vs-verified}).

\paragraph{Baselines}
We compare \method{} against a baseline that we call \baseline{}. \baseline{} runs the same loop with identical settings, including the verification and the acceptance of candidates, but samples all fine-tuning start states from the standard initial-state distribution ($p = 0$ in Eq.~\ref{eq:mixed-init}). The two methods thus differ only in the start states of the fine-tuning episodes, which isolates the contribution of the counterexample start states. In addition, we verify decision trees extracted with VIPER \cite{bastani2018viper} using the same verification settings (Appendix~\ref{sec:app-alternative}).

\subsection{Results}
\label{sec:results}

\begin{figure}[t]
  \centering
  \includegraphics[width=\columnwidth]{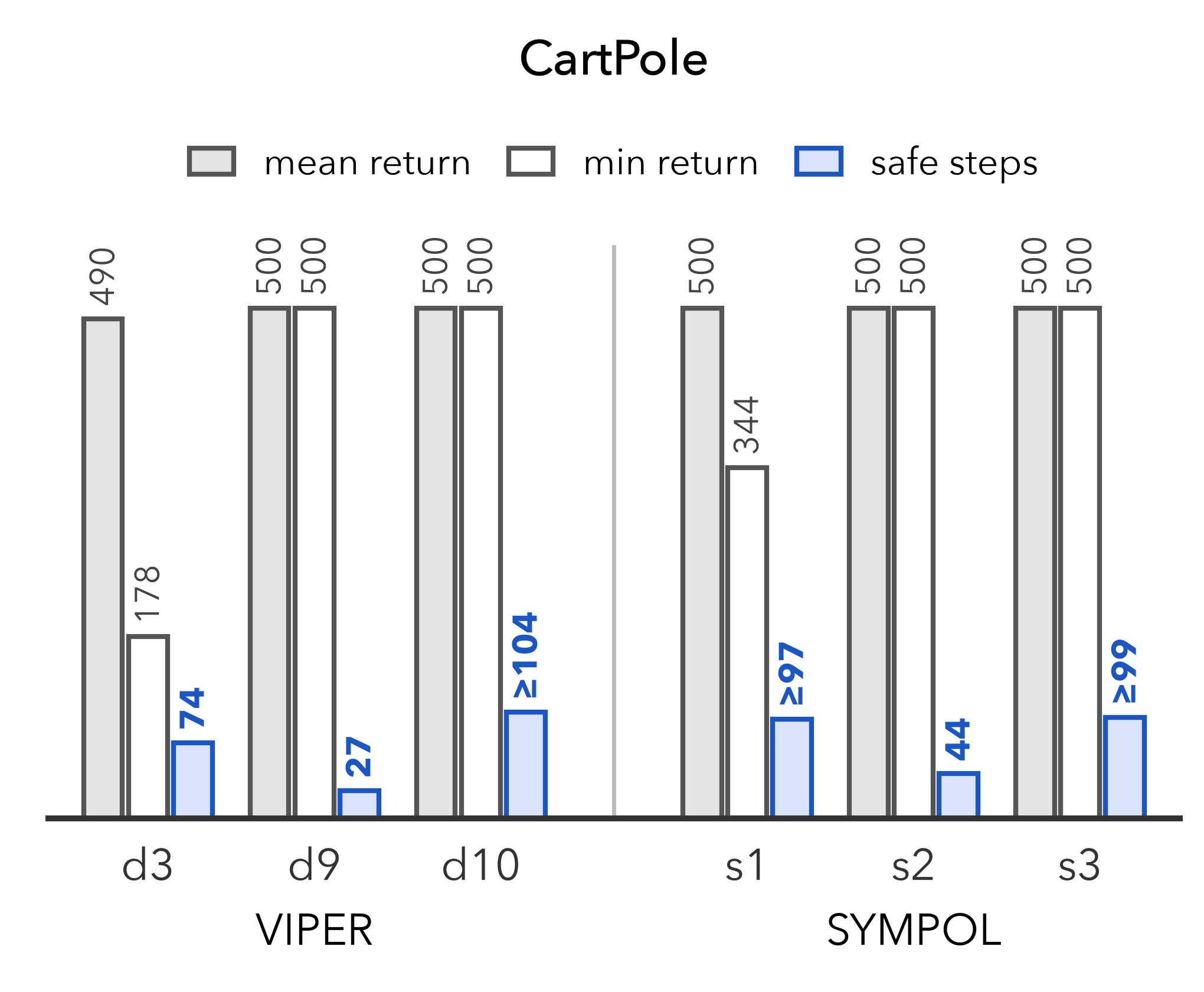}
\caption{High return does not imply safety on CP. Mean and minimum return over 1000 greedy episodes and verified safe steps at $n=200$ for three VIPER trees, labeled by their depth (d3, d9, d10), and three SYMPOL policies of depth 7 (s1--s3). All six were selected by a return $\geq 490$ over 30 episodes. A minimum return of 500 means that no episode violated the constraint. It does not show that the policy is \textit{safe}. $\geq d$: the verifier ran out of memory, so $d$ is a lower bound.}
  \label{fig:return-vs-verified}
\end{figure}

\paragraph{Average return as a performance metric fails for CartPole}
We first trained 21 trees as described by VIPER, and 32 tree ensembles as described by SYMPOL. We then selected 3 of each with high return and verified them with \method{}. The reward issued by the CP environment covers the safety constraint completely: every step is rewarded until an unsafe state is reached. Nevertheless, the average return and the verified safety are decoupled (Fig.~\ref{fig:return-vs-verified}): all six policies obtain a near-maximal average return, yet none is verified \textit{safe} for $n = 200$. The average return therefore neither indicates whether a policy is \textit{safe}, nor does it rank policies by their safety. The reason is that the return averages over sampled start states: the many episodes with high return hide the few start states that lead to an unsafe trajectory. An \textit{unsafe} verdict may be inaccurate (Section~\ref{sec:reach}), but for d3 and d10 we found concrete episodes that violate the safety constraint before step 200 (Appendix~\ref{sec:app-alternative}). For d10, this is the case for only 12 of $10^8$ sampled start states, which an evaluation on 1000 episodes is unlikely to reveal. These violations also remain hidden for a short verification horizon, such as the 10 steps considered by \citet{bastani2018viper}, for which all six policies are verified.

\begin{figure}[t]
  \centering
  \includegraphics[width=\columnwidth]{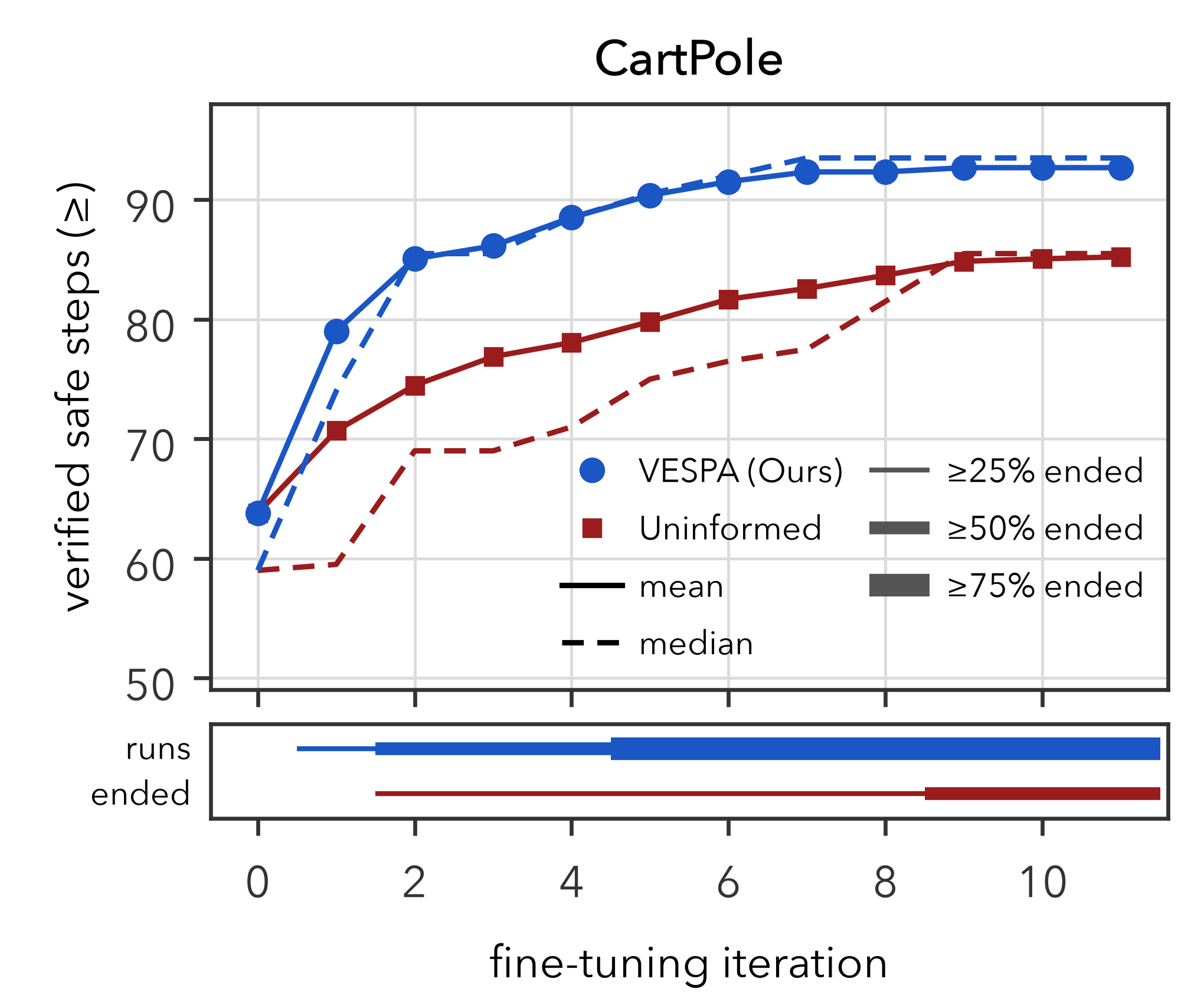}
  \caption{Verified safe steps on CP over the fine-tuning iterations. Mean (solid) and median (dashed) over 48 runs per method (24 input policies $\times$ 2 fine-tuning seeds) of the verified depth of the current policy. Iteration 0 is the input (pretrained) policy. \baseline{} samples all fine-tuning start states from the standard initial-state distribution. A run whose verification runs out of memory keeps its depth as a lower bound ($\geq$), and also ends early (visualized at the bottom).}
  \label{fig:cp-progress}
\end{figure}

\paragraph{\method{} counterexamples represent concrete failures}
A \textit{safe} verdict of \method{} is a proof (Section~\ref{sec:reach}). An \textit{unsafe} verdict, in contrast, is conservative, because the propagated sets can be larger than the exact reachable states. Per step, this over-approximation is small: the median excess of a propagated CP box over its exact image is at most 0.02\% of the image width in every dimension. Over many steps, the excess accumulates, because a set that slightly overlaps a neighboring leaf region is split into an additional piece with a different action. In theory, some counterexamples could therefore be mere artifacts of the over-approximation. However, since the counterexamples clearly improve the fine-tuning (Fig.~\ref{fig:cp-progress}), we conclude that this effect either does not occur or is too small to impair the fine-tuning.

For MC-C, the effect stays small: 98.6\% of the start state regions without a proof contain a concrete episode that misses the goal, so the ratio of safe start state regions is underestimated by at most 1.8 percentage points (Appendix~\ref{sec:app-overapprox}).

\paragraph{\method{} for CP makes considerable progress towards a full safety proof}
Starting from a mean of 63.8 verified safe steps, \method{} verifies a mean of at least 85.1 steps after two fine-tuning iterations, while \baseline{} needs ten iterations to reach the same value (Fig.~\ref{fig:cp-progress}). After the last iteration, the means are at least 92.7 and 85.2 steps, respectively. Most runs end because the verifier runs out of memory, which happens for 40 of 48 \method{} runs and 26 of 48 \baseline{} runs.
The advantage of \method{} shows early. On the same input policy and fine-tuning seed, \method{} is verified for more steps than \baseline{} in 26 of 48 pairs after three fine-tuning iterations and for fewer steps in 5 (Table~\ref{tab:cp} in Appendix~\ref{sec:app-results}). Afterwards, more and more runs reach the memory limit and end with a lower bound, and two lower bounds cannot be ordered. After the last iteration, \method{} is therefore ahead in 14 pairs and behind in 2, while the remaining 32 pairs are equal or cannot be ordered. Because of the memory limit, no run obtains the verdict \textit{safe} for all 200 steps, and the reported gains are lower bounds. As discussed above, \method{} still verifies substantially more steps than prior verifiers for tree policies on CP \cite{bastani2018viper,vasic2022moet}. The deeper proofs come at the cost of return: the mean return of the final policies drops from 496 to 423 for \method{} and to 477 for \baseline{} (medians 492 and 500). A likely reason is that fine-tuning episodes end at step 200, while the return counts 500 steps (Appendix~\ref{sec:app-config}).

\begin{figure}[t]
  \centering
  \includegraphics[width=\columnwidth]{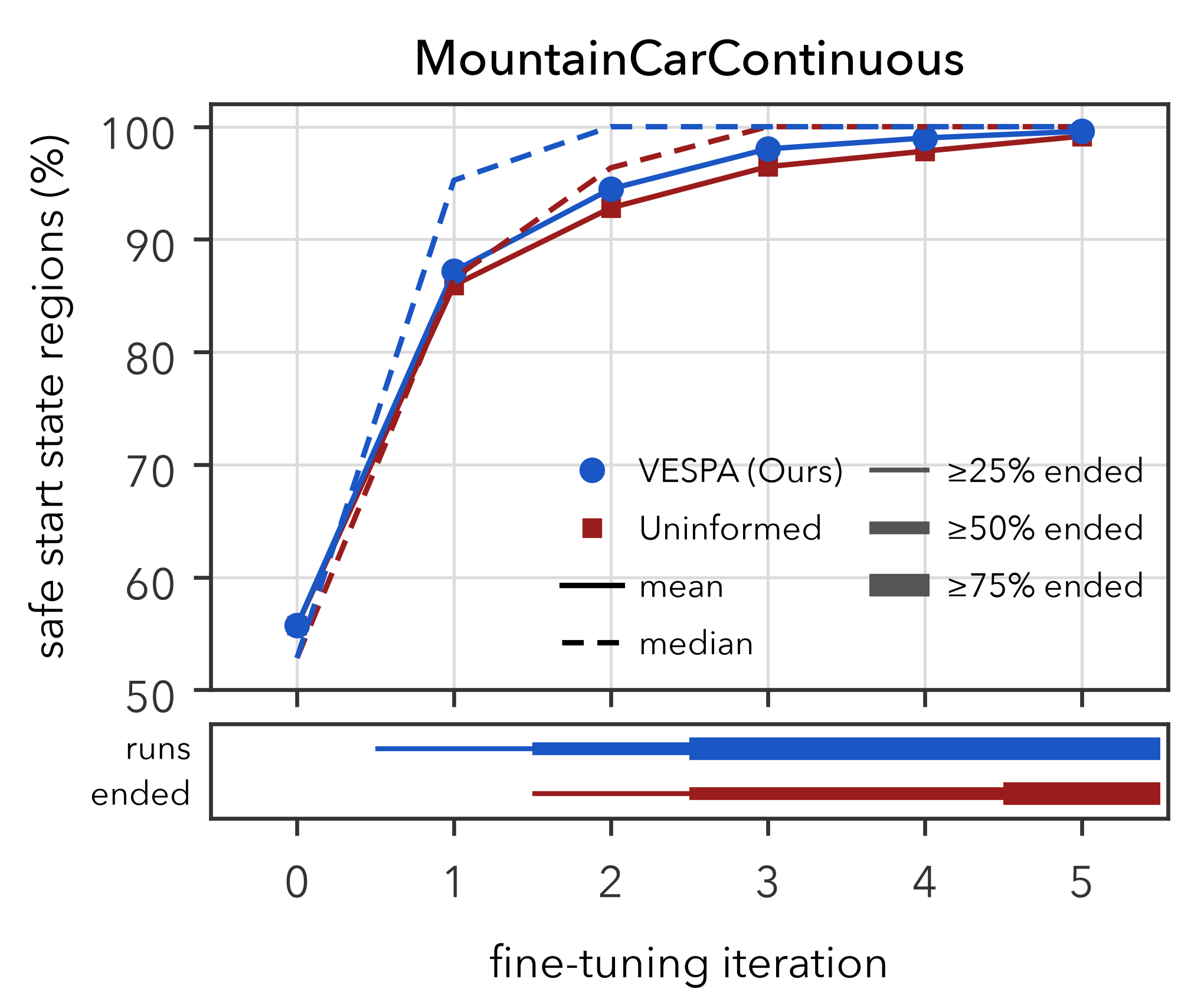}
  \caption{Safe start state regions on MC-C over the fine-tuning iterations. Mean (solid) and median (dashed) over 60 runs per method (12 input policies $\times$ 5 fine-tuning seeds) of the ratio of safe start state regions of the current policy. Iteration 0 is the input (pretrained) policy. A run that is proven \textit{safe} ends early (visualized at the bottom).}
  \label{fig:mcc-progress}
\end{figure}

\paragraph{\method{} for MC-C reliably obtains a safe policy}
None of the 12 input policies for MC-C is \textit{safe}: on average, only 55.7\% of the start state regions provably reach the goal within $n$ steps (Fig.~\ref{fig:mcc-progress}). Within five fine-tuning iterations, \method{} obtains a policy that is proven \textit{safe} on the entire start region $S_0$ in 53 of 60 runs, compared to 46 of 60 runs for \baseline{} (Table~\ref{tab:mcc} in Appendix~\ref{sec:app-results}). In the 42 pairs in which both runs end with a proof, \method{} needs 2.2 fine-tuning iterations until the proof on average and \baseline{} 2.7. In contrast to CP, the proof does not cost return: the mean return increases from 84 to 92 for both methods. Since \textit{n\_lookback} $= n$, all counterexample regions lie in $S_0$. For MC-C, \method{} therefore draws half of the fine-tuning episodes from exactly those start states for which the proof still fails. Whether the input policy is pruned or unpruned has no systematic effect on the results in either environment (Table~\ref{tab:pruning} in Appendix~\ref{sec:app-results}).

\paragraph{Limitations}
The limitations of \method{} are largely those of formal verification in general. A proof refers to a policy and a model of the environment, so the dynamics have to be known, and it covers a bounded horizon $n$. The effort of a reachability analysis grows with the number of state dimensions and the size of the policy. On CP, this is why our proofs end before step 200. They still reach about nine times further than prior work on this benchmark \citep{bastani2018viper, vasic2022moet}, and on MC-C they cover the full horizon.

\section{Related Work}
\label{sec:related}
\paragraph{Verification of Decision Tree Policies}
As discussed in Section~\ref{sec:background}, decision tree policies are well suited for formal verification. For instance, \citet{bastani2018viper} distill deep neural network (DNN) policies into tree policies and then verify them with an SMT solver. They proved correctness of distilled CartPole policies for the first 10 steps. Our verification confirms this, while finding concrete unsafe states at later steps (Section~\ref{sec:results}, Fig.~\ref{fig:return-vs-verified}). Similarly, \citet{jain2024veritas} verify tree policies with a tree-specific solver called \textit{Veritas} and use policy predicate abstraction (PPA) \cite{vinzent2023cegar} to build the reachable state set for an unbounded horizon. \citet{schilling2023dtcs} instead propagate sets of reachable states through tree policies in continuous time. For \method{}, we require the number of verified safe steps and counterexample regions together with the step at which they occur, and thus also choose a set-based representation (Section~\ref{sec:background}).

\paragraph{Verification and Repair of Deep Neural Network Policies}
In contrast to tree ensembles, verifying DNN policies faces additional challenges, especially because their decision boundaries are not axis-aligned. For instance, Trainify \cite{jin2022trainify} discretizes the state space into finitely many abstract states and then trains and verifies a DNN policy on those abstract states. It then refines the discretization based on where the safety constraint is violated and retrains the DNN policy from scratch.

Closest to the adaptation step of \method{} are methods that repair DNN policies with counterexamples from a verifier \cite{boetius2023repair,yang2022repair,wu2024vsrl}. They train on the counterexamples where the verifier finds them, with an added penalty or a loss on verified bounds, whereas \method{} walks the counterexample regions back in time and uses them only as start states for fine-tuning.

\paragraph{Shielding}
As described by \citet{alshiekh2018shielding}, shielding adds a static component that restricts or overrides the actions of the learned policy at runtime. Following the general framework of shielding, \citet{anderson2020revel} propose to alternate between training a DNN policy and a verifiable symbolic policy. Shielding thus shifts the verification problem to the shield, which must itself be provably safe and remain in place at deployment, whereas \method{} verifies and adapts the learned policy itself.

\section{Conclusion and Future Work}
\label{sec:conclusion}
In this paper, we introduced \method{}, a verification and fine-tuning loop that turns counterexamples from reachability analysis into a training signal for fine-tuning. \method{} exploits the hard axis-aligned decision boundaries of tree policies to compute the reachable states, extracts a counterexample region from every violating set, and fine-tunes the policy from these regions. Our evaluation shows that the average return does not indicate whether a tree policy is \textit{safe}, and that counterexamples from the verification are an effective training signal. On CP, \method{} reaches in two iterations the verified safe steps that \baseline{} reaches in ten. On MC-C, it obtains a fully proven policy in 53 of 60 runs.

While \method{} does not yet prove a CP policy \textit{safe} for the full horizon, the limiting factor is the memory required by the verification and not the policy adaptation. Even though our verifier scales to significantly more steps than other approaches, future work could still investigate more memory-efficient state representations. Furthermore, one could extend \method{} to tree ensembles, such as GRANDE \citep{marton2024grande}. We expect that safety-critical applications will ask for a proof before a policy is deployed, and our results show that for tree policies such proofs are within reach.

\section*{Acknowledgments}
This research was supported in part by the German Federal Ministry for Economic Affairs and Climate Action of Germany (BMWK) and by the German Federal Ministry for Research, Technology, and Space (BMFTR).

\bibliography{aaai2027}

\appendix

\section{Detailed Results}
\label{sec:app-results}
Tables~\ref{tab:cp} and~\ref{tab:mcc} report the results per fine-tuning iteration for CP and MC-C, including the average return and the paired comparison of \method{} and \baseline{}.

\begin{table}[!htb]
  \centering
  \small
  \setlength{\tabcolsep}{1.4pt}
  \begin{tabular}{l cc@{\hspace{7pt}} c c@{\hspace{5pt}} c c@{\hspace{7pt}} ccc}
  \toprule
   & \multicolumn{2}{c}{Return} & \multicolumn{4}{c}{$\Delta$ Verified safe steps} & \multicolumn{3}{c}{Pairs} \\
  \cmidrule(l{2pt}r{7pt}){2-3}\cmidrule(l{2pt}r{7pt}){4-7}\cmidrule(l{2pt}r{0pt}){8-10}
  It. & \method{} & Uninf. & \method{} & \#$\geq$ & Uninf. & \#$\geq$ & W & E & L \\
  \midrule
  0 & 496\,{\scriptsize$\pm$9} & 496\,{\scriptsize$\pm$9} & 63.8\,{\scriptsize$\pm$22.3} & 0 & 63.8\,{\scriptsize$\pm$22.3} & 0 & 0 & 48 & 0 \\ \midrule
  1 & 459\,{\scriptsize$\pm$95} & 487\,{\scriptsize$\pm$43} & +15.2\,{\scriptsize$\pm$17.4} & 17 & +6.9\,{\scriptsize$\pm$14.7} & 10 & 18 & 24 & 6 \\
  2 & 445\,{\scriptsize$\pm$100} & 473\,{\scriptsize$\pm$66} & +21.3\,{\scriptsize$\pm$19.9} & 26 & +10.7\,{\scriptsize$\pm$16.5} & 14 & 24 & 17 & 7 \\
  3 & 436\,{\scriptsize$\pm$108} & 467\,{\scriptsize$\pm$75} & +22.4\,{\scriptsize$\pm$19.0} & 30 & +13.1\,{\scriptsize$\pm$17.4} & 15 & 26 & 17 & 5 \\
  4 & 439\,{\scriptsize$\pm$101} & 462\,{\scriptsize$\pm$80} & +24.7\,{\scriptsize$\pm$21.0} & 34 & +14.3\,{\scriptsize$\pm$17.3} & 15 & 25 & 18 & 5 \\
  5 & 429\,{\scriptsize$\pm$105} & 468\,{\scriptsize$\pm$65} & +26.6\,{\scriptsize$\pm$22.0} & 37 & +16.0\,{\scriptsize$\pm$19.3} & 16 & 23 & 22 & 3 \\
  6 & 426\,{\scriptsize$\pm$106} & 474\,{\scriptsize$\pm$59} & +27.7\,{\scriptsize$\pm$23.0} & 39 & +17.9\,{\scriptsize$\pm$19.1} & 18 & 20 & 25 & 3 \\
  7 & 423\,{\scriptsize$\pm$108} & 473\,{\scriptsize$\pm$61} & +28.5\,{\scriptsize$\pm$22.9} & 40 & +18.8\,{\scriptsize$\pm$20.5} & 18 & 21 & 24 & 3 \\
  8 & 423\,{\scriptsize$\pm$108} & 475\,{\scriptsize$\pm$60} & +28.5\,{\scriptsize$\pm$22.9} & 40 & +19.9\,{\scriptsize$\pm$20.5} & 23 & 17 & 28 & 3 \\
  9 & 423\,{\scriptsize$\pm$108} & 477\,{\scriptsize$\pm$58} & +28.9\,{\scriptsize$\pm$22.6} & 40 & +21.1\,{\scriptsize$\pm$20.5} & 25 & 15 & 31 & 2 \\
  10 & 423\,{\scriptsize$\pm$108} & 477\,{\scriptsize$\pm$58} & +28.9\,{\scriptsize$\pm$22.6} & 40 & +21.3\,{\scriptsize$\pm$20.4} & 26 & 14 & 32 & 2 \\
  11 & 423\,{\scriptsize$\pm$108} & 477\,{\scriptsize$\pm$58} & +28.9\,{\scriptsize$\pm$22.6} & 40 & +21.4\,{\scriptsize$\pm$20.5} & 26 & 14 & 32 & 2 \\
  \bottomrule
  \end{tabular}
\caption{CP results over the fine-tuning iterations. 24 input policies $\times$ 2 fine-tuning seeds = 48 paired runs per method, at most 11 iterations. Each value describes the accepted policy after $k$ iterations. Return: mean $\pm$ std of 30 greedy episodes. Verified safe steps: row 0 gives the absolute value of the input policies, rows $k \geq 1$ the change with respect to the run's own input policy (mean $\pm$ std). Runs whose verification ran out of memory enter with their verified depth as a lower bound. \#$\geq$: number of such runs. W/E/L: on the same input policy and seed, \method{} is verified for more steps than \baseline{} (W), for fewer steps (L), or the two are equal or cannot be ordered (E). A lower bound counts as more steps only if it exceeds the exact value of the other run, and two lower bounds cannot be ordered.}
  \label{tab:cp}
\end{table}

\begin{table}[t]
  \centering
  \small
  \setlength{\tabcolsep}{1.4pt}
  \begin{tabular}{l cc@{\hspace{7pt}} c c@{\hspace{5pt}} c c@{\hspace{7pt}} ccc}
  \toprule
   & \multicolumn{2}{c}{Return} & \multicolumn{4}{c}{Safe start state regions (\%)} & \multicolumn{3}{c}{Pairs} \\
  \cmidrule(l{2pt}r{7pt}){2-3}\cmidrule(l{2pt}r{7pt}){4-7}\cmidrule(l{2pt}r{0pt}){8-10}
  It. & \method{} & Uninf. & \method{} & \#safe & Uninf. & \#safe & W & E & L \\
  \midrule
  0 & 84\,{\scriptsize$\pm$11} & 84\,{\scriptsize$\pm$11} & 55.7\,{\scriptsize$\pm$18.0} & 0 & 55.7\,{\scriptsize$\pm$18.0} & 0 & 0 & 60 & 0 \\ \midrule
  1 & 80\,{\scriptsize$\pm$21} & 82\,{\scriptsize$\pm$15} & 87.2\,{\scriptsize$\pm$16.2} & 20 & 85.9\,{\scriptsize$\pm$12.2} & 13 & 28 & 10 & 22 \\
  2 & 89\,{\scriptsize$\pm$9} & 88\,{\scriptsize$\pm$11} & 94.5\,{\scriptsize$\pm$11.8} & 36 & 92.8\,{\scriptsize$\pm$9.7} & 23 & 30 & 16 & 14 \\
  3 & 91\,{\scriptsize$\pm$4} & 90\,{\scriptsize$\pm$7} & 98.0\,{\scriptsize$\pm$6.8} & 45 & 96.5\,{\scriptsize$\pm$6.5} & 32 & 25 & 23 & 12 \\
  4 & 91\,{\scriptsize$\pm$7} & 90\,{\scriptsize$\pm$7} & 99.0\,{\scriptsize$\pm$3.8} & 49 & 97.8\,{\scriptsize$\pm$5.8} & 39 & 20 & 32 & 8 \\
  5 & 92\,{\scriptsize$\pm$2} & 92\,{\scriptsize$\pm$3} & 99.6\,{\scriptsize$\pm$1.6} & 53 & 99.2\,{\scriptsize$\pm$1.9} & 46 & 13 & 42 & 5 \\
  \bottomrule
  \end{tabular}
  \caption{MC-C results over the fine-tuning iterations. 12 input policies $\times$ 5 fine-tuning seeds = 60 paired runs per method, at most 5 iterations of $10^5$ environment steps each. Each value describes the accepted policy after $k$ iterations. A run that is proven safe keeps its policy. Return: mean $\pm$ std of 30 greedy episodes. Safe start state regions: mean $\pm$ std of the ratio of start state regions proven to reach $S_{\mathit{goal}}$ within $n=200$ steps. \#safe: runs proven safe on all of $S_0$. W/E/L: \method{} has a higher, equal or lower ratio than \baseline{} on the same input policy and seed.}
  \label{tab:mcc}
\end{table}

\paragraph{Pruned and Unpruned Input Policies}
Table~\ref{tab:pruning} splits the results of Tables~\ref{tab:cp} and~\ref{tab:mcc} by pruned and unpruned input policies. The number of input safe steps for CP differs slightly, because of rounding errors near the splits, which depend on the tree layout. For \method{}, the two groups behave alike: on CP the mean gain in verified safe steps is $+31.9$ and $+25.9$ at a standard deviation of about 23, and on MC-C 26 and 27 of 30 runs are proven \textit{safe}. \baseline{} varies more. On CP, more of its unpruned runs are verified up to the memory limit (17 vs.\ 9 of 24), whereas on MC-C fewer of its unpruned runs are proven \textit{safe} (21 vs.\ 25 of 30). These two shifts point in opposite directions, and neither is significant after correction for multiple comparisons. We therefore find no systematic effect of pruning.

\begin{table}[t]
  \centering
  \small
  \setlength{\tabcolsep}{1.4pt}
  \begin{tabular}{l cc@{\hspace{7pt}} cc}
  \toprule
   & \multicolumn{2}{c}{Pruned} & \multicolumn{2}{c}{Unpruned} \\
  \cmidrule(l{2pt}r{7pt}){2-3}\cmidrule(l{2pt}r{0pt}){4-5}
   & \method{} & Uninf. & \method{} & Uninf. \\
  \midrule
  \multicolumn{5}{l}{\textit{CP, 24 runs per cell, after 11 iterations}} \\
  Input & \multicolumn{2}{c}{63.5\,{\scriptsize$\pm$22.6}} & \multicolumn{2}{c}{64.1\,{\scriptsize$\pm$22.5}} \\
  $\Delta$ steps & +31.9\,{\scriptsize$\pm$22.5} & +20.6\,{\scriptsize$\pm$22.8} & +25.9\,{\scriptsize$\pm$22.7} & +22.3\,{\scriptsize$\pm$18.4} \\
  \#$\geq$ & 19 & 9 & 21 & 17 \\
  Return & 432\,{\scriptsize$\pm$102} & 462\,{\scriptsize$\pm$77} & 414\,{\scriptsize$\pm$116} & 492\,{\scriptsize$\pm$18} \\
  W/E/L & \multicolumn{2}{c}{10/14/0} & \multicolumn{2}{c}{4/18/2} \\
  \midrule
  \multicolumn{5}{l}{\textit{MC-C, 30 runs per cell, after 5 iterations}} \\
  Input (\%) & \multicolumn{2}{c}{55.7\,{\scriptsize$\pm$18.2}} & \multicolumn{2}{c}{55.7\,{\scriptsize$\pm$18.2}} \\
  Safe reg.\ (\%) & 99.4\,{\scriptsize$\pm$2.1} & 99.5\,{\scriptsize$\pm$1.3} & 99.8\,{\scriptsize$\pm$0.8} & 98.8\,{\scriptsize$\pm$2.3} \\
  \#safe & 26 & 25 & 27 & 21 \\
  Iterations & 74 & 82 & 76 & 111 \\
  Return & 92\,{\scriptsize$\pm$3} & 93\,{\scriptsize$\pm$1} & 92\,{\scriptsize$\pm$2} & 92\,{\scriptsize$\pm$4} \\
  W/E/L & \multicolumn{2}{c}{4/23/3} & \multicolumn{2}{c}{9/19/2} \\
  \bottomrule
  \end{tabular}
\caption{Results split by pruned and unpruned input policies. Every pretraining seed contributes one pruned and one unpruned input policy, and pruning does not alter the policy. CP: 12 seeds $\times$ 2 fine-tuning seeds = 24 runs per cell. MC-C: 6 seeds $\times$ 5 fine-tuning seeds = 30 runs per cell. Input: verified safe steps (CP) or ratio of safe start state regions (MC-C) of the input policies, mean $\pm$ std. All other rows describe the final accepted policy. $\Delta$ steps, \#$\geq$, return and W/E/L for CP are defined as in Table~\ref{tab:cp}. Safe reg., \#safe, return and W/E/L for MC-C are defined as in Table~\ref{tab:mcc}. Iterations: total number of fine-tuning iterations, where a run without a proof counts with all 5.}
  \label{tab:pruning}
\end{table}

\section{Measuring the Over-Approximation}
\label{sec:app-overapprox}
This section describes how we measured the values reported in Section~\ref{sec:results}.

\paragraph{CP, per Step}
We drew 18{,}581 boxes at random from the verifications of 7 input policies and propagated each of them through $\hat{f}$. We compared the result with the smallest box around the exact image, which we approximated from inside by dense sampling, so that the measured excess is an upper bound. Its median lies between 0.001\% and 0.02\% of the width of the exact image, depending on the dimension.

\paragraph{MC-C, over the Full Horizon}
We verified the 6 distinct input policies (the pruned and the unpruned tree of a seed are the same policy) with $2^{10}$ start state regions, a coarser partition than in our experiments, and simulated 1024 concrete episodes from every region. Of the 2{,}737 regions without a proof, 2{,}698 (98.6\%) contain an episode that misses the goal within $n$ steps. Only the remaining 39 regions can be affected by over-approximation. Per policy, these are at most 18 of the 1024 regions, i.e., 1.8 percentage points of the ratio of safe start state regions.

\paragraph{Finer Start Partition}
For 18 pretrained CP policies, we split the part of $S_0$ that leads to the first violation into 16 sub-boxes and verified every sub-box separately. In 16 of 21 cases, all sub-boxes are verified beyond the original first violating step, so the over-approximation does have an influence on the verified safe steps. As discussed in Section~\ref{sec:results}, it does not impair the fine-tuning.

\section{Pretrained Policies}
\label{sec:app-pretrain}
We pretrain a single axis-aligned tree of depth 7 with the SYMPOL configurations \cite{marton2025sympol}, which use PPO \cite{schulman2017ppo}, for 50M environment steps on CP and 1M on MC-C. For CP, we use the checkpoint with the best evaluation return. For MC-C, we keep the best of 5 random trials per seed. The pruning of the SYMPOL implementation removes branches whose split thresholds lie outside the state space and, for discrete actions, merges leaves with the same action.

Table~\ref{tab:policies} lists the number of policies. Tuning policies are used only for the hyperparameter optimization, and all reported results are obtained on the evaluation policies. For CP, the evaluation seeds are those whose initial verification ends with an \textit{unsafe} verdict before it runs out of memory, which is not the case for 13 seeds. Three further seeds were excluded because their pretraining was incomplete or one of their two variants was missing. For MC-C, we only accept a pretrained policy if it reaches the goal within $n$ steps in at least 10\% of the episodes.

\begin{table}[t]
\centering
\small
\setlength{\tabcolsep}{4pt}
\begin{tabular}{lcc}
\toprule
 & CP & MC-C \\
\midrule
Tuning seeds (policies) & 4 (8) & 4 (8) \\
Evaluation seeds (policies) & 12 (24) & 6 (12) \\
Fine-tuning seeds per policy & 2 & 5 \\
Runs per method & 48 & 60 \\
\bottomrule
\end{tabular}
\caption{Number of pretrained policies and runs. Every seed contributes a pruned and an unpruned policy.}
\label{tab:policies}
\end{table}

\section{Verification Settings}
\label{sec:app-verify}
Table~\ref{tab:envs} lists the environments and the verification settings. We verify CP with boxes and MC-C with zonotopes (Section~\ref{sec:reach}), since the MC-C dynamics strongly couple position and velocity, whereas the CP policies switch actions frequently, so that the fragmentation by the tree dominates. Two sets are never merged into a larger enclosing set. The number of generators of a zonotope is limited to 24 by the order reduction of \citet{girard2005zonotopes}, which replaces the smallest generators by an axis-aligned box that encloses them. All sets are propagated in 32-bit floating-point arithmetic with every bound rounded outwards, so that a \textit{safe} verdict is also a proof with respect to floating-point effects. The reward of MC-C favors policies that reach the goal quickly (i.e., far below 200 steps).

\begin{table}[t]
\centering
\small
\setlength{\tabcolsep}{4pt}
\begin{tabular}{@{}lcc@{}}
\toprule
 & CP & MC-C \\
\midrule
Actions & discrete & continuous \\
Episode limit (steps) & 500 & 999 \\
Start region $S_0$ & $[-0.05, 0.05]^4$ & \begin{tabular}{@{}c@{}}$x \in [-0.6, -0.4]$,\\ $v = 0$\end{tabular} \\
Unsafe region $S \setminus S_{\mathit{safe}}$ & \begin{tabular}{@{}c@{}}$|x| > 2.4 \,\lor$\\ $|\theta| > 0.2094$\end{tabular} & -- \\
Goal region $S_{\mathit{goal}}$ & -- & $x \ge 0.45$ \\
Horizon $n$ & 200 & 200 \\
State representation & boxes & zonotopes \\
Start partition & 1 box & $2^{14}$ zonotopes\\
GPU (NVIDIA) & H200 & RTX PRO 6000 \\
Median time per verification & 180\,s & 23\,s\\
\bottomrule
\end{tabular}
\caption{Environments and verification settings. CP follows the CartPole-v1 specification, MC-C the MountainCarContinuous-v0 specification of gymnax \cite{lange2022gymnax}. CP: cart position $x$ and pole angle $\theta$ (rad). MC-C: car position $x$ and velocity $v$.}
\label{tab:envs}
\end{table}

\section{Policy Adaptation and Hyperparameter Optimization}
\label{sec:app-config}

\paragraph{Loop Settings}
A candidate replaces the current policy if it is verified at least as deep (CP), or if it is \textit{safe} or its ratio of safe start state regions is strictly higher (MC-C). A CP candidate whose verification runs out of memory beyond the depth of the current policy is accepted with its verified depth as a lower bound and ends the run. A rejected candidate is discarded, and the current policy is fine-tuned again with a different seed. \method{} and \baseline{} use identical settings and seeds and differ only in $p$ (Eq.~\ref{eq:mixed-init}). \baseline{} still uses the verification, but only as a metric and to accept or reject candidates. On CP, fine-tuning episodes end at the verification horizon $n = 200$, while the reported return uses episodes of up to 500 steps, which can explain the lower return of the fine-tuned CP policies.

\paragraph{Hyperparameters}
Table~\ref{tab:hp} lists the fine-tuning hyperparameters. For CP, they follow \citet{marton2025sympol} with two exceptions: a single learning rate of $3 \cdot 10^{-4}$ for all tree parameters, which is a common default for PPO \cite{schulman2017ppo,andrychowicz2021matters}, and neither weight averaging nor learning rate reduction. A fine-tuning episode on CP ends at the first unsafe state with an additional reward of $-100$ for both methods, whereas pretraining and the fine-tuning on MC-C use the standard reward. We chose \textit{n\_lookback} $= 10$ among the values 2, 6, 10 and 14 in a random search over 25 configurations on the tuning policies. For MC-C, we compared 7 configurations on the tuning policies, separately for \method{} and \baseline{}, and the same configuration ranks first for both. \textit{n\_lookback} $= n = 200$ was fixed, since whether the goal is reached depends heavily on the actions taken at the beginning of an episode.

\begin{table}[t]
\centering
\small
\setlength{\tabcolsep}{4pt}
\begin{tabular}{@{}lcc@{}}
\toprule
Hyperparameter & CP & MC-C \\
\midrule
Learning rate, split thresholds & $3 \cdot 10^{-4}$ & $1.67 \cdot 10^{-4}$ \\
Learning rate, split features & $3 \cdot 10^{-4}$ & $1.46 \cdot 10^{-4}$ \\
Learning rate, leaves & $3 \cdot 10^{-4}$ & $4 \cdot 10^{-3}$ \\
Learning rate, tree weights & $3 \cdot 10^{-4}$ & $1.23 \cdot 10^{-4}$ \\
Learning rate, action log-std & -- & $1 \cdot 10^{-2}$ \\
Upper bound of action log-std at start & -- & $-1$ \\
Learning rate, critic & $1.33 \cdot 10^{-3}$ & $2.06 \cdot 10^{-3}$ \\
Entropy coefficient & 0.2 & 0.1 \\
Discount factor $\gamma$ & 0.99 & 0.999 \\
GAE $\lambda$ & 0.95 & 0.99 \\
Clipping coefficient & 0.1 & 0.1 \\
Value loss coefficient & 0.5 & 0.5 \\
Gradient norm clipping & 1000 & 1000 \\
Parallel environments & 7 & 5 \\
Steps per rollout & 512 & 128 \\
Update epochs & 7 & 8 \\
Minibatch size & 64 & 64 \\
Fine-tuning iterations per run (max.) & 11 & 5 \\
Environment steps per iteration & $10^5$ & $10^5$ \\
Reward at the first unsafe state & $-100$ & -- \\
\textit{n\_lookback} & 10 & 200 \\
Mixing probability $p$ & \multicolumn{2}{c}{0.5 (\method{}), 0 (Uninf.)} \\
\bottomrule
\end{tabular}
\caption{Fine-tuning hyperparameters of \method{} and \baseline{}.}
\label{tab:hp}
\end{table}

\section{Verification of Alternative Pretraining Methods}
\label{sec:app-alternative}
\method{} does not make any assumptions on how the decision tree policy was obtained. In addition to SYMPOL, we therefore verified 21 decision trees extracted with VIPER \cite{bastani2018viper} from a DQN teacher (maximum leaf counts of 2, 4, 8, 16, 32, 64 and 128 with 3 seeds each), using the settings of Appendix~\ref{sec:app-verify}. Fig.~\ref{fig:return-vs-verified} shows the three trees with an average return of at least 490 over 30 episodes. None of the 21 trees is verified \textit{safe} for 200 steps. For 20 of them, we found concrete episodes that violate the safety constraint before step 200 among $10^8$ sampled start states.

Tree ensemble policies are future work, with Gradient-Based Decision Tree Ensembles (GRANDE) \citep{marton2024grande} being the natural extension of SYMPOL trees.

\end{document}